\documentclass[graybox]{svmult}

\usepackage{type1cm}
\usepackage{newtxtext}
\usepackage{newtxmath}

\usepackage{makeidx}
\usepackage{graphicx}
\usepackage{multicol}
\usepackage[bottom]{footmisc}

\usepackage{amsmath}
\usepackage[numbers,sort&compress]{natbib}
\usepackage{booktabs}
\usepackage{hyperref}

\makeindex

\begin{document}

\title*{On The Role of K-Space Acquisition in MRI Reconstruction Domain-Generalization}
\titlerunning{K-Space Acquisition in MRI Domain-Generalization}

\author{Mohammed Wattad, Tamir Shor, and Alex Bronstein}
\authorrunning{M. Wattad et al.}

\institute{Mohammed Wattad \at Faculty of Computer Science, Technion, Israel, \email{Mohammed-wa@campus.technion.ac.il}
\and Tamir Shor \at Faculty of Computer Science, Technion, Israel, \email{tamir.shor@campus.technion.ac.il}
\and Alex Bronstein \at Faculty of Computer Science, Technion, Israel, \email{bron@cs.technion.ac.il}}

\maketitle

\abstract{Recent work has established learned k-space acquisition patterns as a promising direction for improving reconstruction quality in accelerated Magnetic Resonance Imaging (MRI). Despite encouraging results, most existing research focuses on acquisition patterns optimized for a single dataset or modality, with limited consideration of their transferability across imaging domains. In this work, we demonstrate that the benefits of learned k-space sampling can extend beyond the training domain, enabling superior reconstruction performance under domain shifts. Our study presents two main contributions. First, through systematic evaluation across datasets and acquisition paradigms, we show that models trained with learned sampling patterns exhibit improved generalization under cross-domain settings. Second, we propose a novel method that enhances domain robustness by introducing acquisition uncertainty during training—stochastically perturbing k-space trajectories to simulate variability across scanners and imaging conditions. Our results highlight the importance of treating k-space trajectory design not merely as an acceleration mechanism, but as an active degree of freedom for improving domain generalization in MRI reconstruction.}

\footnote{Code for reproducing all experiments is available at \href{https://github.com/mohammedwttd/On-The-Role-of-K-Space-Acquisition-in-MRI-Reconstruction-Domain-Generalization}{\textcolor{purple}{https://github.com/mohammedwttd/On-The-Role-of-K-Space-Acquisition-in-MRI-Reconstruction-Domain-Generalization}}}
\section{Introduction}
\label{sec:intro}

Magnetic Resonance Imaging (MRI) is a commonly-preferred medical-imaging technique, mostly due to its noninvasive nature, exceptional soft-tissue contrast, and absence of ionizing radiation. Despite these advantages, MRI suffers from inherently long acquisition times, often leading to limited applicability and increased patient discomfort at scan time.
These inhibitions have fostered growing research focus dedicated to MR acceleration. This acceleration is typically achieved by undersampling the k-space prior to algorithmic reconstruction to filter sub-Nyquist sampling artifacts \cite{zaitsev2015motion}.

Classical compressed sensing (CS) methods \cite{lustig2007sparse} demonstrated that accurate reconstruction is possible from far fewer samples when appropriate priors (e.g., sparsity or total variation) are imposed. However, CS-based optimization often requires long iterative solvers and cannot fully exploit the vast data now available from modern MRI systems. The reconstruction task remains an ill-posed inverse problem—estimating the fully-sampled image from undersampled measurements.

The emergence of deep learning has transformed MRI reconstruction by learning powerful data-driven priors directly from MR data \cite{schlemper2018deep,hammernik2018learning,fastmri_dataset}. Large-scale datasets like fastMRI \cite{fastmri_dataset} have further standardized evaluation and accelerated community progress. Specifically, many previous works demonstrate the potential of directly optimizing k-space acquisition patterns alongside the reconstruction module, in order to get favorable reconstruction quality and acceleration factors \cite{learning_optimal_kspace,shor2023multi,active_kspace_rl, weiss2019joint}.

In spite of promising results reported for learned MR-acquisition in existing research, most related previous work demonstrates the advantage of these approaches when shaping acquisition patterns during training for reconstruction within the same MR dataset and modality as the one seen during test-time. While ample research had been devoted for efficient and differentiable modeling of k-space sampling patterns \cite{weiss2021pilot,wang2021bjork,learning_optimal_kspace}, the exploration of these sampling patterns for purposes beyond in-domain reconstruction remains limited.

Our key distinction in this work is that the additional degrees-of-freedom tunable for a model that jointly optimizes reconstruction \textit{and} k-space acquisition parameters can be utilized to mitigate performance drops increased in the case of MR-domain shifts. Importantly, as we demonstrate in this study, this utilization is not bound to model that actually use learned k-space acquisition patterns at inference time.

Our contribution is two-fold:
\begin{enumerate}
\item We perform a systematic evaluation of the susceptibility of various MR reconstruction approaches to MR-domain shift. We compare fixed and learned sampling patterns, and find that learned k-space sampling is significantly favorable in tested domain shift scenarios.
\item We develop a novel method for promoting cross-domain shift robustness, based on introducing both stochastic and adversarial acquisition uncertainty during training. Our method portrays favorable performance compared to counterparts reliant on the introduction of noise in image-domain. Importantly, the applicability of our method does not depend on whether k-space sampling patterns are optimized towards reconstruction during training.
\end{enumerate}

\section{Background}
\label{sec:background}

\subsection{Learned MRI Acquisition}
\label{subsec:acq_recap}
Great volume of research has been devoted to the exploration of fixed and learned k-space undersampling patterns. Traditional handcrafted sampling patterns such as Cartesian (uniform or variable-density) grids \cite{lustig2007sparse}, radial trajectories \cite{block2007undersampled}, and spiral trajectories \cite{glover1999spiral} have been extensively explored, each offering distinct trade-offs between hardware efficiency, artifact structure, and reconstruction robustness.

Undersampling strategies aim to accelerate scans by reducing the number of acquired measurements while relying on reconstruction algorithms \cite{lustig2007sparse,ronneberger2015unet, du2021adaptive,shor2024team} to fill in the missing information. These reconstruction algorithms are tasked with denoising sub-Nyquist sampling artifacts \cite{zaitsev2015motion} originated from sparse undersampling. The distribution of these artifacts is governed by the chosen undersampling scheme, rendering the choice of undersampling patterns crucial for efficient MR acceleration.

Conventional fixed trajectories (e.g., Cartesian, radial, spiral) are designed with heuristic principles, such as prioritizing low-frequency regions that contribute most to image contrast. However, they are not optimized for a given reconstruction model and can limit achievable quality at high acceleration factors.
Learned trajectories, in contrast, jointly optimize the sampling pattern along with the reconstruction network parameters. \cite{weiss2019joint} optimize Cartesian acquisition patterns based on a quantile-based binarization of a continuous, learned sampling mask. PILOT \cite{weiss2021pilot} directly parameterize a set of sampling points interpolated by B-SPLINE interpolation to learn kinematically-constrained, non-Cartesian sampling patterns. BJORK \cite{wang2021bjork} learn non-Cartesian sampling masks modeled as quadratic B-spline kernels, enforcing gradient continuity for hardware feasibility.

\subsection{Domain Shift and the Role of Trajectory}
A key obstacle in deploying deep-learning-based MRI reconstruction is \textbf{domain shift}: models trained on one distribution may perform poorly when evaluated on another \cite{zhou2021domain_generalization}. In MRI, this can occur across imaging protocols, hardware, organs-of-interests, or imaging planes (e.g. axial, sagittal, coronal).

Existing research has attempted to address domain shift primarily through image- or feature-level modifications. Methods such as \textbf{self-supervised learning (SSL)} \cite{yaman2020ssdu} and \textbf{domain generalization (DG)} \cite{zhou2021domain_generalization} focus on learning domain-invariant \textit{features} or \textit{representations} from the reconstructed images or the latent space of the network. Similarly, distributionally robust optimization (DRO) techniques \cite{robust_mri_recon} aim to ensure performance across the worst-case training subset, implicitly leading to broader generalization. However, these approaches still often require access to either target-domain data for adaptation, or strong assumptions about the relationship between source and target distributions, and they fundamentally treat the acquisition process as a fixed, external factor.

In this work, we hypothesize that the acquisition trajectory itself can serve as a powerful and under-utilized lever for mitigating domain shift. Prior efforts in trajectory optimization focused only on finding the single \textit{best} trajectory for a \textit{fixed} training distribution \cite{learning_optimal_kspace, pilot, weiss2019joint}. We propose a different paradigm: using the trajectory as a regularization mechanism during training. By introducing controlled \emph{trajectory perturbations}—stochastic variations in the k-space sampling process—we simulate the real-world variability and noise encountered across different scanners and acquisition settings. This approach moves beyond purely image- or feature-level regularization by directly addressing the physics of data acquisition. It encourages the model to become less reliant on domain-specific sampling patterns and more robust to physical variability, opening the possibility of using trajectory design not only for acceleration but as a mitigation mechanism for cross-domain degradation in reconstruction performance.

\section{Methodology}
\label{sec:method}

\subsection{Overview and Experimental Design}
The core of our methodology is the evaluation of \textbf{trajectory-aware regularization} as a mechanism to enhance the robustness and domain generalization of accelerated Magnetic Resonance Imaging (MRI) reconstruction networks. We investigate the effect of three distinct domain generalization (DG) strategies on four fundamental model baselines, and then assess their performance under cross-domain shift.

\textbf{Domain Generalization (DG) Strategies}
We compare a standard clean training setup (No DG) against three methods that introduce targeted variability into the training process:
\begin{enumerate}
    \item \textbf{Random Trajectory Noise:} Simulates realistic, small, random acquisition imperfections.
    \item \textbf{Adversarial Trajectory Noise:} Models the worst-case, structured acquisition deviations to test robustness limits.
    \item \textbf{Image Noise:} Serves as a classical, image-domain augmentation baseline for comparison.
\end{enumerate}

\noindent\textbf{Baseline Architectures and Trajectory Settings.}
We define four model baselines, combining two architectures with two sampling paradigms:
U-Net (convolutional) and PT-ViT-L (transformer), each trained with either a fixed or learned trajectory.

By applying the three DG strategies to all four baseline variants, we create a total of $4 \times (1 \text{(Clean)} + 3 \text{(DG)}) = 16$ experimental models, allowing for a thorough assessment of the relative effect of trajectory-based regularization on domain generalization.

\subsection{Clean Training Setup and Acquisition Modeling}

We first establish the standard, clean training procedure and elaborate on the architectures and acquisition modeling used across all experiments.

\subsubsection{Reconstruction Architectures}

\paragraph{U-Net \cite{ronneberger2015unet}.}
The U-Net serves as our primary convolutional baseline , mostly since it is one of the most prevalent reconstruction models in MRI acceleration literature. The U-Net utilizes an encoder-decoder structure with skip connections, which is highly effective for capturing multi-scale image features. For clean training, the U-Net is trained end-to-end to minimize the reconstruction loss.

\paragraph{PT-ViT-L \cite{vision_transformers_mri}.}
The PT-ViT-L (Pre-trained Vision Transformer, Large) is our attention-based baseline, designed to capture long-range spatial dependencies. We use this model due to its SOTA performance, and the fact it represents attention-based models, which are also widely use in medical imaging and general computer-vision. Crucially, the model from \cite{vision_transformers_mri} is initialized with weights pre-trained on the ImageNet, and subsequently fine-tuned on the fastMRI dataset. The inclusion of this model allows us to assess how trajectory-aware methods interact with strong, globally-attentive priors.

\subsubsection{Acquisition and Trajectory Modeling}

\paragraph{Acquisition Overview.}
The acquisition process differs between the Cartesian and Radial sampling strategies. In the \textbf{Cartesian acquisition}, data are sampled on a uniform Fourier grid, where each \textit{k}-space point corresponds to a regularly spaced location along both frequency-encoding and phase-encoding directions. In contrast, the \textbf{Radial acquisition} employs non-uniform Fourier sampling, where each trajectory is defined by a list of off-grid points forming radial spokes that pass through the center of \textit{k}-space. These trajectories are subsequently used to reconstruct images via the non-uniform Fast Fourier Transform (NUFFT).

We evaluate models under two primary k-space sampling paradigms:

\begin{enumerate}
\item \textbf{Fixed Trajectory:} This baseline uses a predefined, non-learnable sampling mask, such as a \textbf{Cartesian mask} (e.g., variable density or uniform) or a \textbf{Radial spoke pattern}. The forward operator $\mathcal{F}_u$ is fixed, and only the reconstruction network $f_\theta$ is optimized.

\item \textbf{Learned Trajectory:} In this setting, the sampling pattern $u$ is jointly optimized with the reconstruction network parameters $\theta$. The objective is formulated as:
\begin{equation}
\min_{u, \theta} \; \| x_{\text{gt}} - f_\theta(\mathcal{F}_u^{-1} y) \|_1 + \mathcal{R}(\theta),
\end{equation}
where $y = \mathcal{F}_u x_{\text{gt}}$ are the undersampled measurements, $\mathcal{F}_u^{-1}$ is the inverse acquisition operator, and $\mathcal{R}(\theta)$ is a regularization term (e.g., weight decay) applied only to $\theta$. For our modeling of k-space acquisition parameters we follow PILOT \cite{weiss2021pilot} — the set of acquisition parameters is a fixed number of k-space sampling coordinates. The undersampling mask is attained for this set of coordinates by B-SPLINE interpolation and Non-uniform FFT \cite{dutt1993fast}.
\end{enumerate}

\begin{figure}[t]
    \centering
    \begin{minipage}[b]{0.32\textwidth}
        \includegraphics[width=\textwidth]{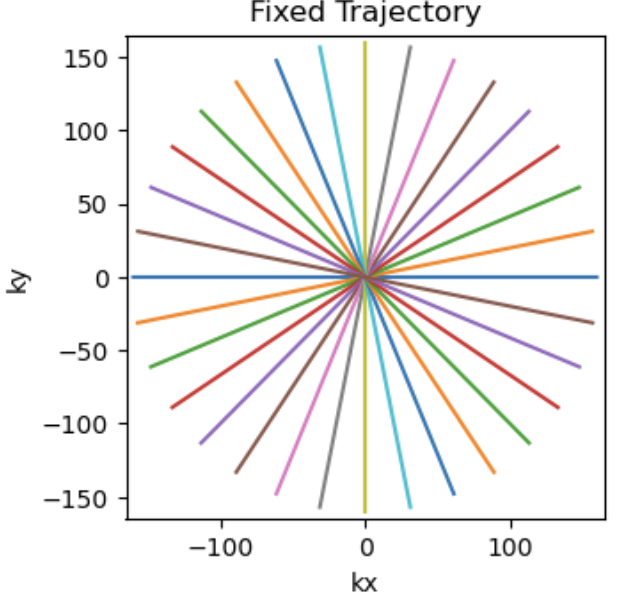}
        \caption{Fixed Radial Trajectory.}
        \label{fig:fixed_radial}
    \end{minipage}
    \hfill
    \begin{minipage}[b]{0.64\textwidth}
        \includegraphics[width=\textwidth]{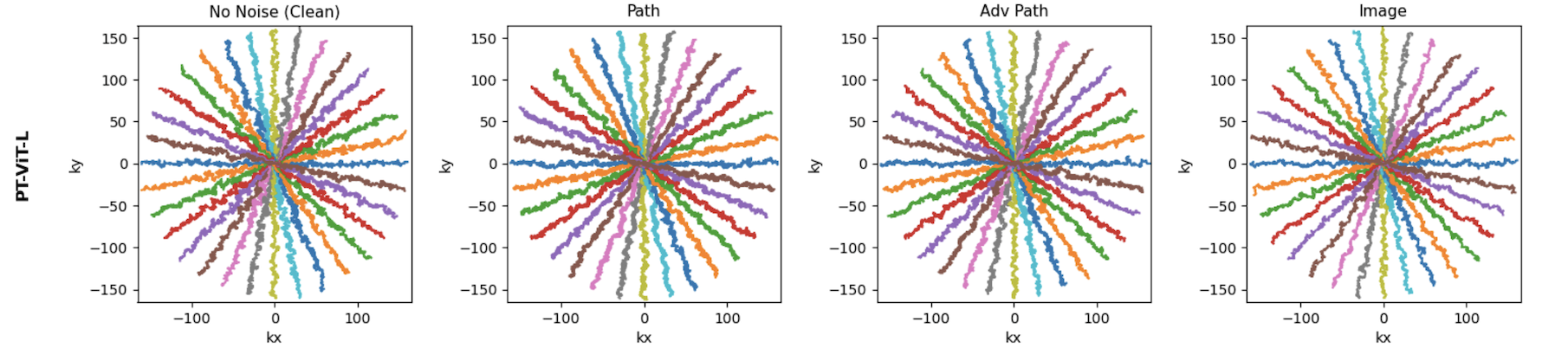}
        \caption{Learned Radial Trajectories (PT-ViT-L).}
        \label{fig:learned_radial}
    \end{minipage}
\end{figure}

For radial trajectories (Fig.~\ref{fig:fixed_radial}), we employ a \emph{Radial Interpolation Gap} to stabilize training: we optimize a sparser set of points and reconstruct the full trajectory via scheduled linear interpolation (decreasing the interpolation gap $g$ linearly over half the total epochs), allowing the model to learn coarse features before fine-tuning the full resolution.

\subsection{Domain Generalization (DG) Methods}

We introduce three noise-injection variants to improve robustness to domain shift. For all variants, the reconstruction loss is
$\mathcal{L}(\theta) = \| x_{\text{gt}} - f_\theta(y') \|_1$,
where the noisy measurement $y'$ is produced with probability $p \sim \text{Bernoulli}(0.5)$ using the schedule in Section~\ref{sched_st}.

\subsubsection{Random Trajectory Noise}
This method aims to simulate the minor, random hardware and patient motion artifacts that introduce $\mathbf{real-world}$ $\mathbf{acquisition}$ $\mathbf{variability}$. By exposing the model to slightly jittered k-space coordinates, we promote the learning of features that are less sensitive to precise sampling placement.

\paragraph{Noise Model.}
\begin{itemize}
    \item \textbf{Cartesian Sampling:} Only the sampled lines ($u_i=1$) are perturbed independently using an integer-valued Gaussian offset $\delta_i \sim \mathcal{N}_{\text{int}}(0, \tau^2)$, with $\tau = 10$. The final trajectory $u'$ ensures all perturbed positions are unique.
    \item \textbf{Radial Sampling:} Each point in the spoke trajectory is perturbed by a Gaussian offset $\delta \sim \mathcal{N}(0, \tau^2)$, with $\tau = 30$, applied to both $x$ and $y$ coordinates.
\end{itemize}
The noisy measurements are $y' = \mathcal{F}_{u'} x + n$, where $n$ is measurement noise.

\subsubsection{Adversarial Trajectory Noise}
This method explores the $\mathbf{upper}$ $\mathbf{bound}$ $\mathbf{of}$ $\mathbf{trajectory}$ $\mathbf{variability}$ by intentionally finding the perturbations that maximize the reconstruction error. Training the network to overcome these "worst-case" deviations ensures maximum robustness.

\paragraph{Optimization Formulation (Radial).}
For the radial trajectory $u$, the adversarial perturbation $\delta$ is computed by solving a constrained inner maximization problem before the standard outer minimization (reconstruction):
$\max_{\|\delta\|_\infty \le \epsilon} \mathcal{L}(\theta, u + \delta)$.
It is approximated using a single-step Fast Gradient Sign Method (FGSM) \cite{goodfellow2015explaining} with an $L_\infty$ constraint $\epsilon$, where
$\delta = \epsilon \cdot \text{sign}(\nabla_u \mathcal{L})$ and $u' = u + \delta$.
For Cartesian sampling, adversarial noise is injected using a bitwise XOR operation: a perturbation mask $\lambda$ flips $2 \cdot \text{numBits} = 8$ entries based on high (for $u=1$) and low (for $u=0$) loss gradient magnitudes, simulating a worst-case line mis-sampling.

It is important to note that our adversarial formulation and noise models (both adversarial and random) are not intended to perfectly simulate a specific physical scanner malfunction (e.g. eddy currents). Rather, it serves solely as a robust optimization mechanism. By identifying and training against the "worst-case" trajectory deviation via FGSM, we minimize the model's tendency to overfit to the specific acquisition patterns during training. This promotes the stability of the reconstruction network, even when subject to the milder, non-adversarial trajectory imperfections (e.g., gradient delays) encountered in real-world cross-domain scenarios.

\subsubsection{Image Noise}
We use image-domain noise as a standard augmentation to improve robustness to general image corruption (e.g., intensity fluctuations or sensor noise). This serves as a $\mathbf{non-acquisition-specific}$ $\mathbf{baseline}$. Gaussian noise $\epsilon \sim \mathcal{N}(0, \sigma^2)$ with $\sigma = 6 \times 10^{-5}$ is added to the ground truth $x$: $x' = x + \epsilon$. The noisy image is then converted back to k-space to obtain $y' = \mathcal{F}_u x' + n$.

\begin{figure}[t]
    \centering
    \begin{minipage}[b]{0.48\textwidth}
        \centering
        \includegraphics[width=\textwidth]{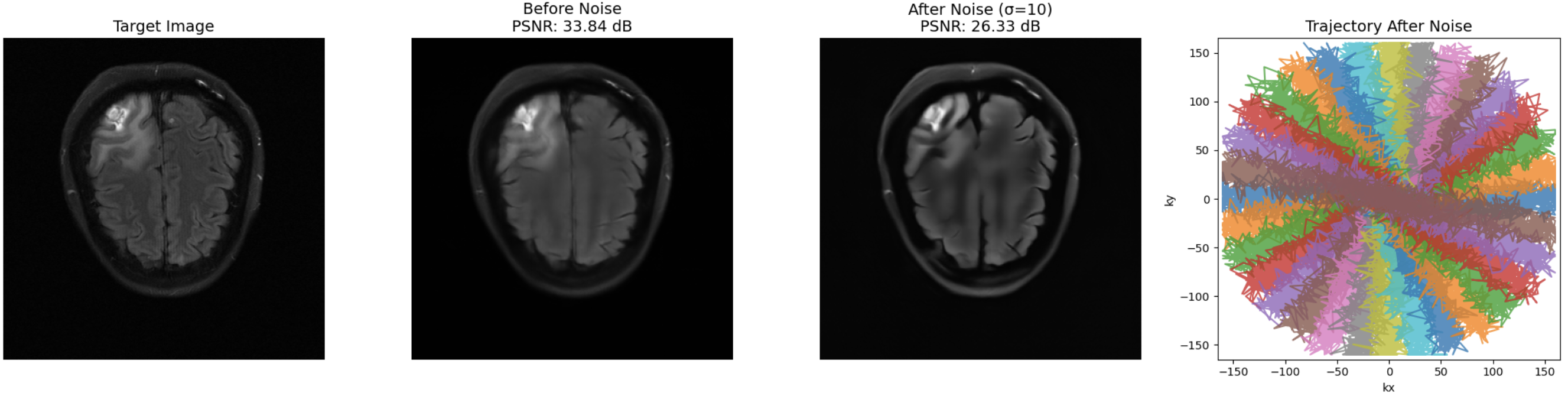}
        \caption{Standard traj. noise (PT-ViT-L).}
        \label{fig:traj_noise_standard}
    \end{minipage}
    \hfill
    \begin{minipage}[b]{0.48\textwidth}
        \centering
        \includegraphics[width=\textwidth]{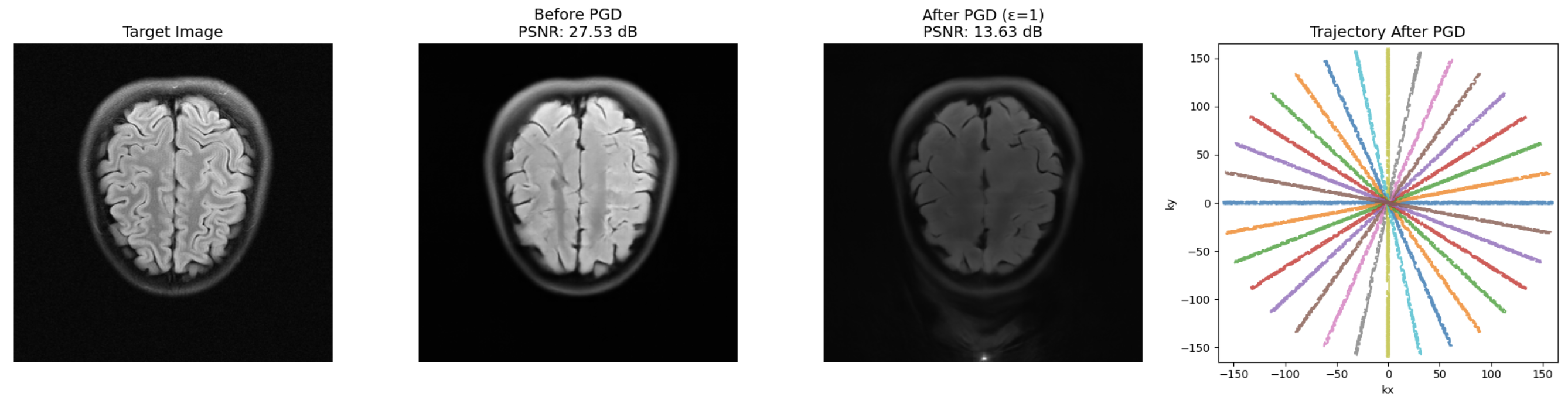}
        \caption{Adversarial traj. noise (PT-ViT-L).}
        \label{fig:traj_noise_adv}
    \end{minipage}
\end{figure}

\subsection{Noise Scheduling}
\label{sched_st}
To maximize training stability and robustness, all noise types ($\mathbf{Trajectory}$ $\mathbf{Noise}$, $\mathbf{Adv.}$ $\mathbf{Trajectory}$ $\mathbf{Noise}$, and $\mathbf{Image}$ $\mathbf{Noise}$) are introduced using a scheduled noise intensity $\epsilon(t)$ that follows a warmup-then-decay strategy:
\begin{equation}
\epsilon(t) =
\begin{cases}
\epsilon_\text{max} \frac{t}{t_\text{warmup}}, & 0 \le t \le t_\text{warmup} \\
\epsilon_\text{max} \left( 1 - \frac{t - t_\text{warmup}}{T - t_\text{warmup}} \right), & t_\text{warmup} < t \le T
\end{cases}
\end{equation}
where $t$ is the current epoch, $T$ is the total number of epochs, and $t_\text{warmup} = 4$. This schedule ensures the model gradually adapts to noise and then consolidates its learning as the noise level is tapered off.

\subsection{Training Procedure Details}

The optimization schedule for all models follows the same linear warmup-then-decay strategy used for noise injection. \emph{In all experiments, a mini-batch size of $B=1$ was used}.

\subsubsection{Noise Parameter Selection and Calibration}

Noise parameters for all DG methods were tuned to balance two goals. The noise must be \textit{strong enough} to cause noticeable degradation when applied to clean-trained models, confirming a meaningful distribution shift, yet \textit{moderate enough} to preserve strong in-domain performance on fastMRI. Specifically, we used $\tau{=}10$ (Cartesian) and $\tau{=}30$ (Radial) for \emph{Trajectory Noise}, $\epsilon{=}1$ and $numBits{=}4$ for \emph{Adversarial Trajectory Noise}, and $\sigma{=}6{\times}10^{-5}$ for \emph{Image Noise}. This calibration ensures realistic robustness evaluation while maintaining fair cross-method comparison.

\begin{table}[t]
\caption{In-domain fastMRI PSNR (Mean $\pm$ SD) for Cartesian (top) and Radial (bottom) trajectories.}
\label{tab:indomain_all}
\centering
\begin{tabular}{p{1.4cm}c cccc}
\svhline\noalign{\smallskip}
\textbf{Model} & \textbf{TL} & \textbf{No Noise} & \textbf{Image} & \textbf{Trajectory} & \textbf{Traj. Adv} \\
\noalign{\smallskip}\svhline\noalign{\smallskip}
\textit{Cartesian} & & & & & \\
PT-ViT-L & Yes & \textbf{32.65} $\pm$ 4.98 & 32.27 $\pm$ 4.91 & 31.91 $\pm$ 4.77 & 32.26 $\pm$ 4.86 \\
PT-ViT-L & No  & 31.88 $\pm$ 4.85 & 31.82 $\pm$ 4.86 & 31.83 $\pm$ 4.80 & 31.90 $\pm$ 4.82 \\
U-Net    & Yes & \textbf{32.03} $\pm$ 4.79 & 31.81 $\pm$ 4.76 & 31.49 $\pm$ 4.71 & 31.59 $\pm$ 4.79 \\
U-Net    & No  & 31.55 $\pm$ 4.71 & 31.30 $\pm$ 4.69 & 31.07 $\pm$ 4.72 & 31.23 $\pm$ 4.65 \\
\noalign{\smallskip}\hline\noalign{\smallskip}
\textit{Radial} & & & & & \\
PT-ViT-L & Yes & 31.57 $\pm$ 4.72 & 31.02 $\pm$ 4.58 & 30.77 $\pm$ 4.53 & 30.82 $\pm$ 4.75 \\
PT-ViT-L & No  & 29.56 $\pm$ 4.60 & 29.51 $\pm$ 4.55 & 29.42 $\pm$ 4.53 & 29.41 $\pm$ 4.54 \\
U-Net    & Yes & \textbf{33.10} $\pm$ 5.06 & 32.35 $\pm$ 4.93 & 31.36 $\pm$ 4.71 & 32.08 $\pm$ 5.05 \\
U-Net    & No  & 27.95 $\pm$ 4.29 & 27.62 $\pm$ 4.30 & 27.58 $\pm$ 4.18 & 27.75 $\pm$ 4.31 \\
\noalign{\smallskip}\svhline\noalign{\smallskip}
\end{tabular}
\end{table}

\paragraph{U-Net Training.}
The training data for the U-Net consisted of approximately 6000 examples. The U-Net training phase consisted of 40 epochs. The maximum reconstruction learning rate was set to $5\times10^{-4}$. Subsampling learning rates for trajectory optimization were set to $0.025$ for Cartesian and $0.005$ for radial.

\paragraph{PT-ViT-L Training.}
The fine tuning data for the PT-ViT-L consisted of approximately 3000 examples. The pretraining stage was executed cleanly, without noise. Fine-tuning was performed for 30 epochs with a maximum reconstruction learning rate of $1\times10^{-4}$. Subsampling learning rates were set to $0.5$ for Cartesian and $0.01$ for radial, consistent with prior trajectory-learning literature.

\paragraph{Acquisition Parameters.}
For 4$\times$ accelerated Cartesian sampling, 80 out of 320 phase-encoding lines are acquired (1 = acquired, 0 = skipped). For 4$\times$ accelerated Radial sampling, 16 shots of 1600 points each yield a total of 25,600 sampled points.

\section{Results}
\label{sec:results}

\paragraph{Data and Evaluation Setup.}
Both the M4Raw \cite{lyu2023m4raw} and fastMRI \cite{fastmri_dataset} brain datasets consist of multi-coil acquisitions. For each slice, the multi-coil k-space data are first transformed to the image domain using a 2D inverse Fourier transform. For M4Raw, the resulting complex-valued images are interpolated from $256 \times 256$ to $320 \times 320$, while fastMRI brain images are originally $320 \times 320$. The multi-coil images are then collapsed into a single complex-valued image using the root-sum-of-squares (RSS) method. These images serve as input to the sampling model, and after the acquisition simulation, the resulting real-valued input images are normalized to the range $[0,1]$ before being fed to the reconstruction network. The fully sampled target images, also at $320 \times 320$, are normalized to $[0,1]$ and serve as ground-truth supervision. For evaluation, 2000 samples from each dataset were used. All models were trained and evaluated on an NVIDIA RTX A6000 GPU.

\subsection{Quality Metrics}
We use \textit{paired distribution analysis} to account for sample-wise variability. For models A and B, the PSNR difference per sample $i$ is
$\Delta \text{PSNR}_i = \text{PSNR}_i^\text{A} - \text{PSNR}_i^\text{B}$.
The mean and standard deviation of $\Delta \text{PSNR}$ quantify the average relative performance and its consistency: positive mean favors Model A, and lower std indicates more consistent improvement.

\subsection{In-Domain Performance}
We report each method’s performance on the training source domain (fastMRI), serving as a baseline for later domain-shift evaluation.
Table \ref{tab:indomain_all} shows the average in-domain PSNR on the fastMRI test set.

\paragraph{Observation 1: In-Domain Trade-off.} As a form of regularization, all noise strategies (\emph{Image}, \emph{Trajectory}, \emph{Adv}) result in a slight decrease in peak in-domain PSNR (Table \ref{tab:indomain_all}) compared to the respective clean-trained (\emph{No Noise}) baseline. The highest fidelity on the source domain is consistently achieved by the clean models with \emph{Trajectory Learning (TL)} enabled. Our results show that learned acquisition patterns outperform similar models with fixed patterns, in alignment with trends from previous work \cite{weiss2021pilot,wang2021bjork,shor2023multi}.

\begin{table}[t]
\caption{Cross-Domain Paired PSNR differences (No TL - With TL) on M4Raw.}
\label{tab:paired_m4raw}
\centering
\begin{tabular}{l l l r r}
\svhline\noalign{\smallskip}
\textbf{Mask} & \textbf{Model} & \textbf{Noise Type} & \textbf{Mean Diff} & \textbf{Std Dev} \\
\noalign{\smallskip}\svhline\noalign{\smallskip}
Cart. & PT-ViT & None      & -0.675 & 0.780 \\
      &        & Trajectory      & 0.083  & 0.690 \\
      &        & Traj Adv.  & -0.128 & 0.673 \\
      &        & Image     & -0.613 & 0.679 \\
\noalign{\smallskip}
Cart. & U-Net  & None      & -0.511 & 0.848 \\
      &        & Trajectory      & -0.229 & 0.526 \\
      &        & Traj Adv.  & 0.279  & 0.699 \\
      &        & Image     & -0.979 & 0.813 \\
\noalign{\smallskip}\hline\noalign{\smallskip}
Rad.  & PT-ViT & None      & -1.651 & 1.345 \\
      &        & Trajectory      & -1.247 & 1.049 \\
      &        & Traj Adv.  & -1.281 & 1.109 \\
      &        & Image     & -0.629 & 1.079 \\
\noalign{\smallskip}
Rad.  & U-Net  & None      & -4.376 & 1.392 \\
      &        & Trajectory      & -2.838 & 1.030 \\
      &        & Traj Adv.  & -3.338 & 1.237 \\
      &        & Image     & -4.165 & 1.512 \\
\noalign{\smallskip}\svhline\noalign{\smallskip}
\end{tabular}
\end{table}

\begin{figure}[t]
\sidecaption
\includegraphics[width=0.6\linewidth]{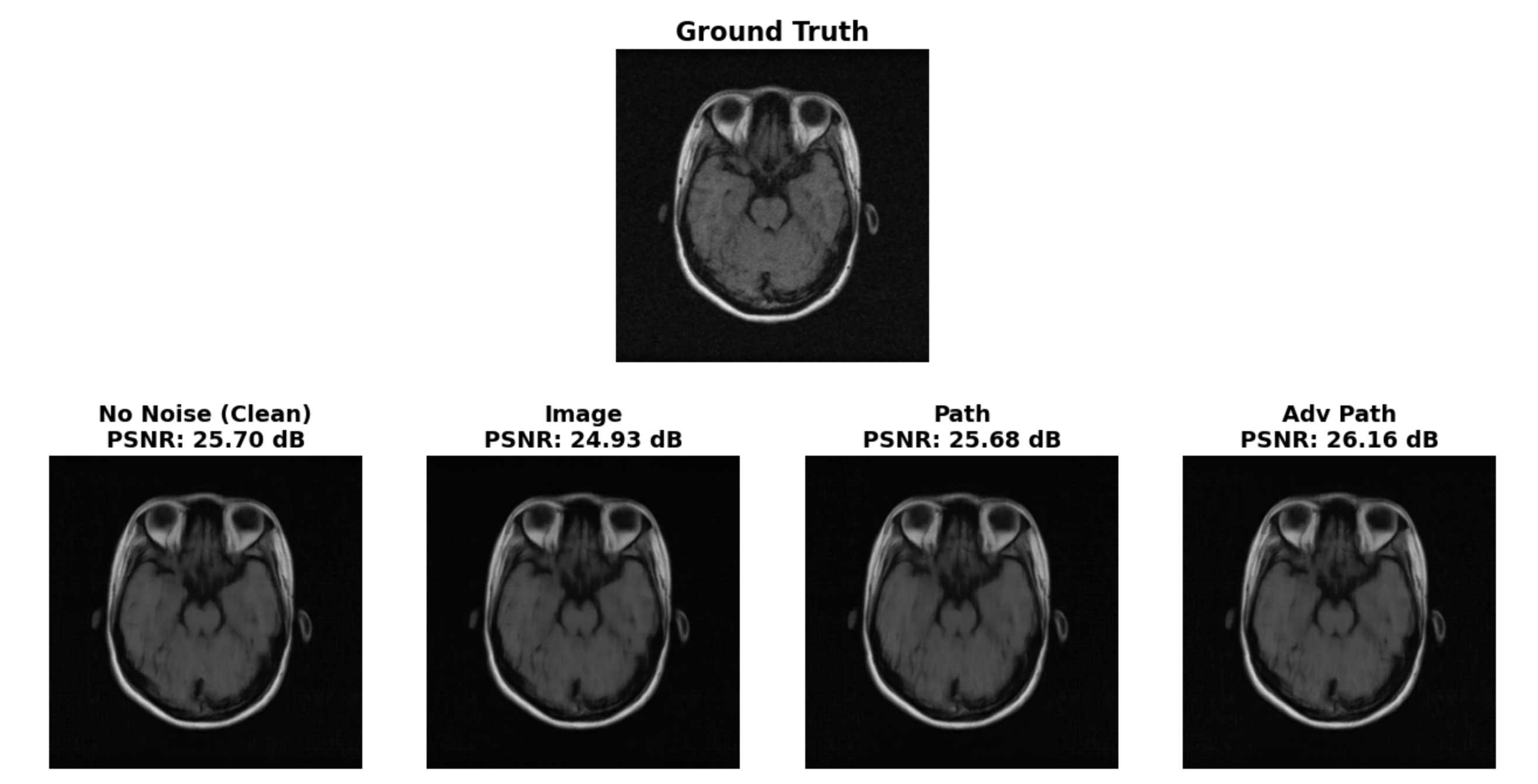}
\caption{\textbf{PT-ViT-L} reconstructions without TL on Cartesian; trajectory noise outperforms image noise.}
\label{fig:ptvit_noise_comparison}
\end{figure}

\subsection{Cross-Domain Analysis I: Impact of Trajectory Learning (TL)}
To assess the domain-generalization (DG) benefit of trajectory learning, we compute the paired difference: $\Delta \text{PSNR} = \text{PSNR}_{\text{No TL}} - \text{PSNR}_{\text{TL}}$ (i.e., No TL minus With TL) on the M4Raw dataset. A \textbf{negative mean difference} implies that \emph{Trajectory Learning improves generalization}.

\begin{figure}[t]
\sidecaption
\includegraphics[width=0.6\linewidth]{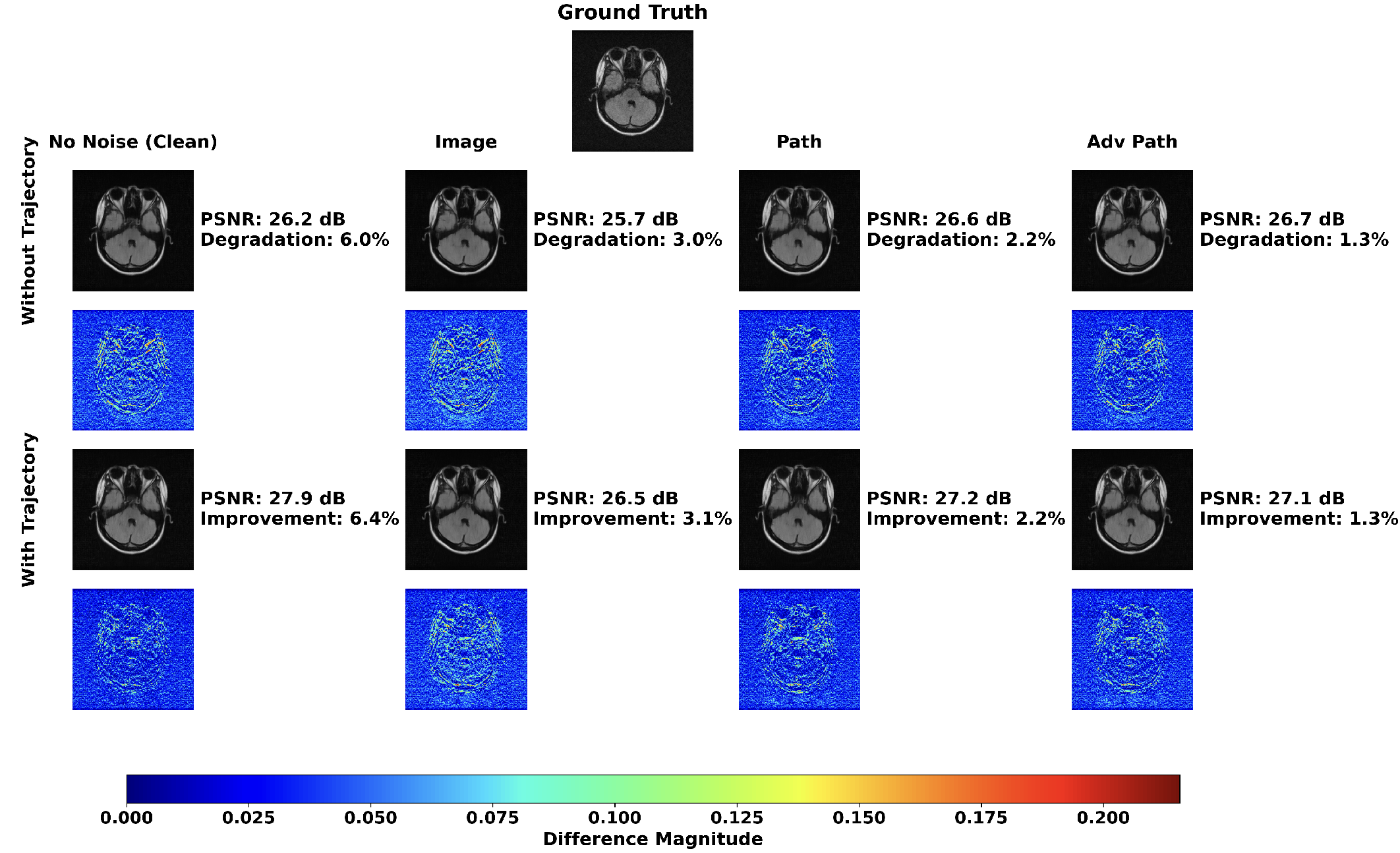}
\caption{Example of \textbf{PT-ViT-L} on Cartesian M4Raw. Each pair shows models with/without TL. Imp. is $\frac{\text{PSNR}_{\text{TL}} - \text{PSNR}_{\text{No-TL}}}{\text{PSNR}_{\text{No-TL}}} \!\times\! 100\%$.}
\label{fig:ptvit_tl_comparison}
\end{figure}

\begin{figure}[t]
\sidecaption
\includegraphics[width=0.6\linewidth]{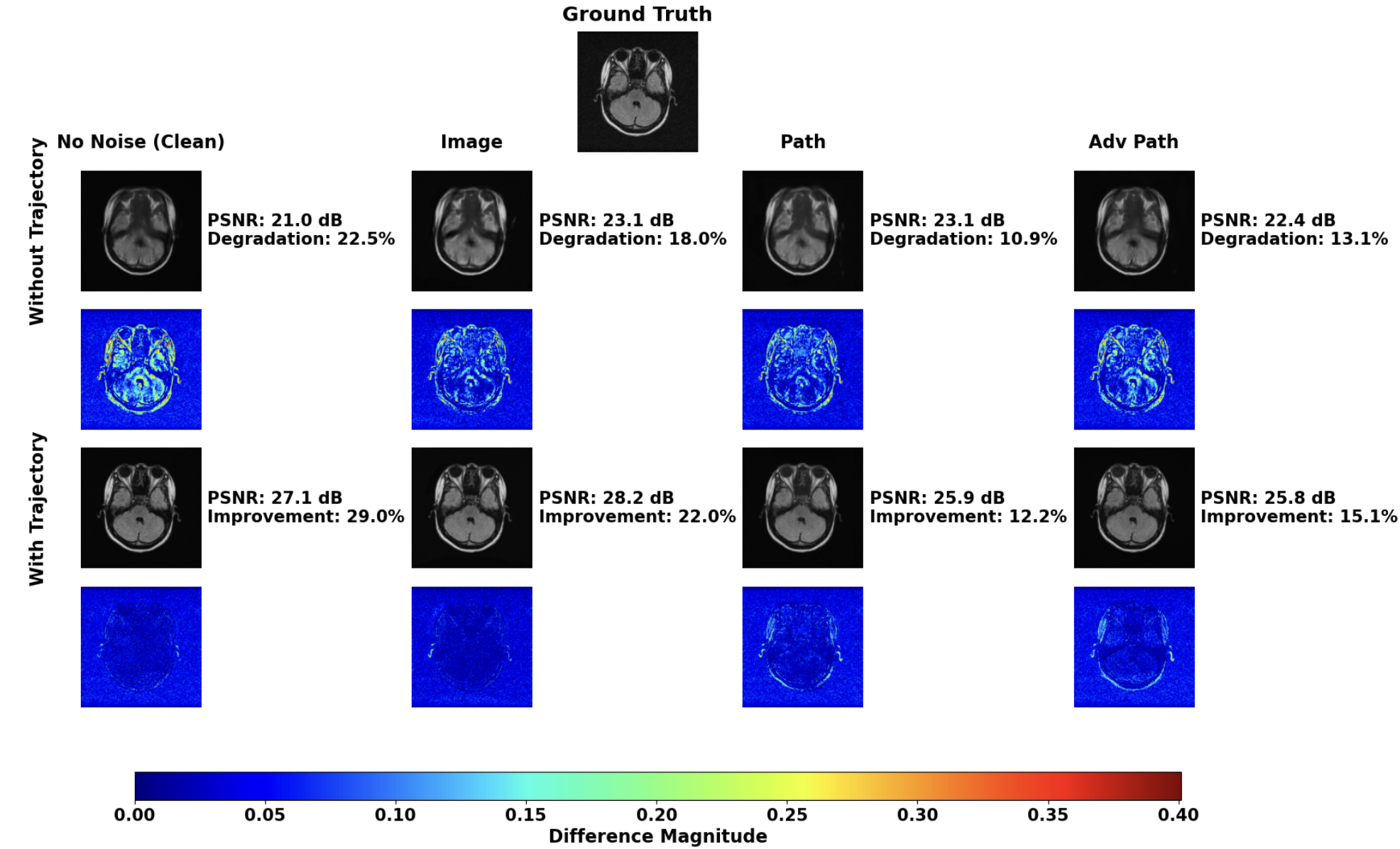}
\caption{Example of \textbf{U-Net} on Radial M4Raw across four noise types. Each pair shows models with and without TL. Imp. is $\frac{\text{PSNR}_{\text{TL}} - \text{PSNR}_{\text{No-TL}}}{\text{PSNR}_{\text{No-TL}}} \!\times\! 100\%$.}
\label{fig:unet_tl_comparison}
\end{figure}

\paragraph{Observation 2: TL Allows Favorable Performance Degradation Under Domain-Shift.}
As shown in Table \ref{tab:paired_m4raw}, \emph{Trajectory Learning provides a dominant, consistent generalization benefit} across almost all configurations ($\Delta \text{PSNR}$ are mostly strongly negative). For radial acquisition the improvement is largest for U-Net models, achieving up to $\mathbf{4.376}$ dB PSNR gain over the fixed radial pattern baseline. Figure \ref{fig:unet_tl_comparison} visually demonstrates this substantial benefit for U-Net on radial data, where TL models show improvements close to $\sim$30\% (PSNR) with visually apparent reconstruction quality gains. For Cartesian sampling the trends are mostly similar, however U-Net with Adversarial Trajectory Noise seems to have favorable results without trajectory learning. We attribute this outcome to the small number of trainable parameters of a U-Net without trajectory learning, that is much lower than all other baselines.
This suggests that for Cartesian sampling, the robustness achieved by trajectory noise can sometimes outweigh the specialization benefit of a learned pattern.
These trends are also evident from figure \ref{fig:ptvit_tl_comparison}, highlighting how TL models improve over their No-TL peers for a specific example.

\paragraph{Observation 3: Sampling-Pattern Uncertainty Out-favors Image-Domain Uncertainty for DG.} In most test-cases, the introduction of uncertainty in sampling-domain (i.e. noise types "Trajectory", "Trajectory Adv.") during training decreases domain-shift reconstruction quality degradation, in comparison with the uncertainty introduced in image-domain ("Image" noise type) or with the case where no DG training techniques are applied ("None" noise type). This conclusion is further supported by observations  4 and 5 ahead. We believe these findings can be explained by the fact that sampling trajectory noise is more closely-related to the distributions changes observed during domain-shifts originated from varying sampling settings. Namely, even if reconstruction models operate over data in image-domain, MR domain shifts mostly occur in k-space domain.

\subsection{Cross-Domain Analysis II: Efficacy of Noise on Fixed Trajectories (No-TL)}

This analysis focuses only on models trained with a \emph{fixed, handcrafted sampling trajectory (No-TL)}. We compare noise strategies against the clean-trained (\emph{No Noise}, \emph{No-TL}) baseline model. The paired difference is: $\Delta \text{PSNR} = \text{PSNR}_{\text{Noise}} - \text{PSNR}_{\text{No Noise}}$. A \textbf{positive mean difference} indicates the noise strategy improves DG performance.

\paragraph{Observation 4: Trajectory Noise Drives Superior Cartesian Generalization.}
Table \ref{tab:fixed_noise_m4raw} further-support conclusions drawn in observation 3.
As seen in Table \ref{tab:fixed_noise_m4raw} (top), for fixed Cartesian acquisition,
\emph{trajectory noise is significantly more effective than image noise} at improving domain generalization.
\emph{Adversarial Trajectory Noise} yields the highest positive gain for both PT-ViT-L (+0.176) and U-Net (+0.101).
In sharp contrast, \emph{Image Noise} strongly \emph{harms} generalization (up to -1.499 for U-Net), confirming that simulating physical misalignments (trajectory noise) is critical for building robust Cartesian models.
This behavior is visually illustrated in Figure~\ref{fig:ptvit_noise_comparison}.

\begin{table}[t]
\caption{Cross-Domain PSNR Improvement for Fixed Trajectory Models (M4Raw).}
\label{tab:fixed_noise_m4raw}
\centering
\begin{tabular}{l l l r r}
\svhline\noalign{\smallskip}
\textbf{Mask} & \textbf{Model} & \textbf{Noise Type} & \textbf{Mean Diff} & \textbf{Std Dev} \\
\noalign{\smallskip}\svhline\noalign{\smallskip}
Cart. & PT-ViT & Adv. Trajectory & \textbf{0.176} & 0.343 \\
      &        & Trajectory    & 0.036 & 0.432 \\
      &        & Image    & -0.358 & 0.504 \\
\noalign{\smallskip}
Cart. & U-Net  & Adv. Trajectory & \textbf{0.101} & 1.090 \\
      &        & Trajectory    & -0.464 & 0.939 \\
      &        & Image    & -1.499 & 0.797 \\
\noalign{\smallskip}\hline\noalign{\smallskip}
Rad. & PT-ViT & Trajectory    & \textbf{0.012} & 0.727 \\
     &        & Image    & -0.126 & 0.759 \\
     &        & Adv. Trajectory & -0.147 & 0.694 \\
\noalign{\smallskip}
Rad. & U-Net  & Image    & \textbf{0.579} & 1.196 \\
     &        & Trajectory    & 0.569 & 0.730 \\
     &        & Adv. Trajectory & 0.408 & 0.677 \\
\noalign{\smallskip}\svhline\noalign{\smallskip}
\end{tabular}
\end{table}

\paragraph{Observation 5: Trajectory Noise is the Most Consistent Radial DG Strategy.} For the U-Net model with fixed radial acquisition (Table \ref{tab:fixed_noise_m4raw}, bottom), both \emph{Image Noise} and \emph{Trajectory Noise} improve generalization. However, \emph{Random Trajectory Noise} (+0.569 PSNR gain) is achieved with significantly lower standard deviation ($\sigma=0.730$) compared to \emph{Image Noise} ($\sigma=1.196$), suggesting \emph{Trajectory Noise} is a more reliable and consistent method for improving robustness to radial shifts. The robust PT-ViT-L sees only negligible changes from any noise type.

\section{Discussion}
\label{sec:discussion}

\subsection{Conclusion}
This work introduced a trajectory-aware perspective for MRI reconstruction \emph{Domain Generalization (DG)}, demonstrating that k-space sampling patterns can be applied to enhance cross-dataset robustness without requiring target-domain supervision. Our systematic analysis across U-Net and PT-ViT-L backbones, paired with Cartesian and Radial acquisition, yielded three key findings:
\begin{enumerate}
    \item \textbf{Trajectory learning is beneficial for domain generalization:} Following observation 2, jointly learning the trajectory with the reconstruction network provided the slightest performance degradation in cross-domain PSNR, demonstrating that optimized sampling pattern can promote domain-transferability.
    \item \textbf{Trajectory noise (stochastic or adversarial) mostly out-favors image-domain noise in DG robust training:} For fixed Cartesian and radial acquisition, both \emph{Random Trajectory Noise} and \emph{Adversarial Trajectory Noise} consistently provided positive PSNR gains, while traditional \emph{Image Noise} was detrimental (up to $\mathbf{-1.499}$ PSNR loss for U-Net). This trend is apparent in most test cases for learned sampling schemes as well.

\end{enumerate}

The robustness observed with trajectory noise can be attributed to its operation in the Fourier domain. Unlike Gaussian noise introduced in image-domain, which is spatially independent, perturbations in k-space coordinates introduce structured, global artifacts in the image domain (e.g., aliasing and blurring) that mimic the distribution of physical acquisition errors. By regularizing the reconstruction model to resolve these specific frequency-domain inconsistencies during training, the model learns invariance to phase errors and non-ideal sampling densities—phenomena that vary significantly during domain shifts.

Our results show that on top of promoting enhanced acceleration, k-space trajectories can pose as a a regularizer for diminishing domain overfitting. Thus, incorporating trajectory awareness—through learning or perturbation—can make reconstruction models more adaptable and reliable.

\subsection{Limitations and Future Work}
In spite of notable findings, we wish to denote several limitations of our work. First, our work is focused on DG across datasets, making our findings mostly relevant for domain-shifts introduced by differences in scan configurations. We leave the exploration of other common MR domain shifts (e.g. imaged organs, imaging plane) to future work. These limitations are mostly originated from current scarcity of diverse MR datasets containing full raw k-space data.

While our evaluation focuses on brain MRI, the physical principles governing k-space acquisition are consistent across anatomies, suggesting that the domain-generalization benefits observed here are likely transferrable to other modalities (e.g., knee or abdomen), though quantifying this precise gain is left for future work.
Additionally, we emphasize that our work addresses the Domain Generalization (DG) setting, where the target domain is unseen during training. A comparison to domain-specific fine-tuning (Domain Adaptation) would be inequitable, as fine-tuning assumes access to target-domain data (and often ground truth labels), which is a major constraint to be overcome in zero-shot deployment scenarios.

Future work can extend our findings by integrating \emph{domain adversarial objectives} or \emph{meta-learning frameworks} with trajectory optimization to further improve cross-domain generalization or adaptation.

\end{document}